# A New Deep Boosted CNN and Ensemble Learning based IoT Malware Detection

Saddam Hussain Khan[1] *, Wasi Ullah[1]

[1]Department of Computer Systems Engineering, University of Engineering and Applied Science (UEAS), Swat, Pakistan;

**Email Address:** saddamhkhan@ueas.edu.pk

## Abstract

Security issues are threatened in various types of networks, especially in the Internet of Things (IoT) environment that requires early detection. IoT is the network of real-time devices like home automation systems and can be controlled by open-source android devices, which can be an open ground for attackers. Attackers can access the network credentials, initiate a different kind of security breach, and compromises network control. Therefore, timely detecting the increasing number of sophisticated malware attacks is the challenge to ensure the credibility of network protection. In this regard, we have developed a new malware detection framework, Deep Squeezed-Boosted and Ensemble Learning (DSBEL), comprised of novel Squeezed-Boosted Boundary-Region Split-Transform-Merge (SB-BR-STM) CNN and ensemble learning. The proposed STM block employs multi-path dilated convolutional, Boundary, and regional operations to capture the homogenous and heterogeneous global malicious patterns. Moreover, diverse feature maps are achieved using transfer learning and multi-path-based squeezing and boosting at initial and final levels to learn minute pattern variations. Finally, the boosted discriminative features are extracted from the developed deep SB-BR-STM CNN and provided to the ensemble classifiers (SVM, MLP., and AdabooSTM1) to improve the hybrid learning generalization. The performance analysis of the proposed DSBEL framework and SB-BR-STM CNN against the existing techniques have been evaluated by the IOT_Malware dataset on standard performance measures. Evaluation results show progressive performance as 98.50% accuracy, 97.12% F1-Score, 91.91% MCC, 95.97 % Recall, and 98.42 % Precision. The proposed malware analysis framework is robust and helpful for the timely detection of malicious activity and suggests future strategies.

# 1 Introduction

Malware is an undesired software that can harm digital devices like computers, android, and especially the Internet of Things (IoT) devices. IoT has gained popularity expeditiously in the digital market due to its robust features and applications. The IoT devices improved human life quality and will increase to 43 billion in 2023. The concept of IoT is to transform real objects into virtual objects having unique addresses and can be driven by the popular open-source android devices. In this emerging technology, intelligent devices share their information and resources accordingly [1]. Several vital roles perform the new web of interconnected devices in our daily lives, like smart health care, home automation, intelligent education environment, and industry. There are many widespread Applications in various fields like monitoring agriculture soil state [2], e-health and healthcare applications [3–6], and deployment of intelligent communication devices on battlefields for military application [7,8]. To build up a supply chain link between industry and end-users, industry 4.0 exploited this new concept [9]. Industrial IoT (IIoT) has undoubtedly contributed well towards products and innovations in improving industrial infrastructure.

IoT devices are heterogeneous in both structures and network protocols, where each heterogeneous device has a unique microprocessor characteristic [10]. So, this is the major cause that the IoT industry is lagging in security protocols and becoming enlarged attack surface, leading to security breaches. This provides tunnels for cyber criminals to exploit the vulnerabilities and utilize the attacks for their illegal actions. IoT devices are vulnerable to security attacks, easily exploited, and compromise network control. Recently, more than 178 million IoT devices, like webcams, medical devices, routers, etc., have been exposed to attackers because new technology is the key entry point [11]. Therefore, it is highly desired to secure IoT devices, and security countermeasures are required to protect them from cyberattacks.

Major cyber security concerns include malware attacks, DDoS, botnets, rootkits, intrusions, ransomware, and compromise nodes. Malware is software that includes viruses, adware, Trojan horses, spyware, etc., and can harm computers and web devices. In a malware attack, the attacker can gain access to the network and take complete control without any awareness. It is becoming a massive barrier to malware analysts and making the ground interest for security researchers. Compared to other digital devices, there is no regular patching in IoT devices because of their embedded nature [10], and it is impossible to implement the security protocols on all IoT devices uniformly. These security breaches are interpreted in Figure 1. Android malware detection reached 26.61 million in 2018 and noticed a 520,000 monthly increase [12]. Therefore, there may be a mechanism for detecting malware attacks under these issues in IoT devices to take immediate action and secure the system or device before compromising.

IoT Malware analysis comes under the umbrella of static and dynamic analysis. The static malware detection method is the way of detection by signature-based, permission-based, and bytecode-based methods. However, static malware analysis is simple and can be easily fooled by obfuscation, and runtime vulnerabilities lead unnoticed. On the other side, the dynamic method is the way of detection in which the applications are executed in isolated platforms such as (simulators, sandbox, and virtual machines). The environment is secured, trusted, and undetectable, tracking their behaviors during the execution of the suspicious file, whether normal or malicious. The traditional detection techniques are relied on built-in signature libraries and mainly on human involvement, and it is hard to detect malware grown very extensively [13,14]. Moreover, the malware binary files have been converted into an image where bytes are mapped into the pixels, and Machine learning (ML) methods have been employed to detect ELF-based malware [15]. However, ML methods required additional effort for feature extraction from images to get domain expert knowledge for malware detection. Lately, deep learning (DL) and deep CNN models have been considered for IoT malware detection [16,17].

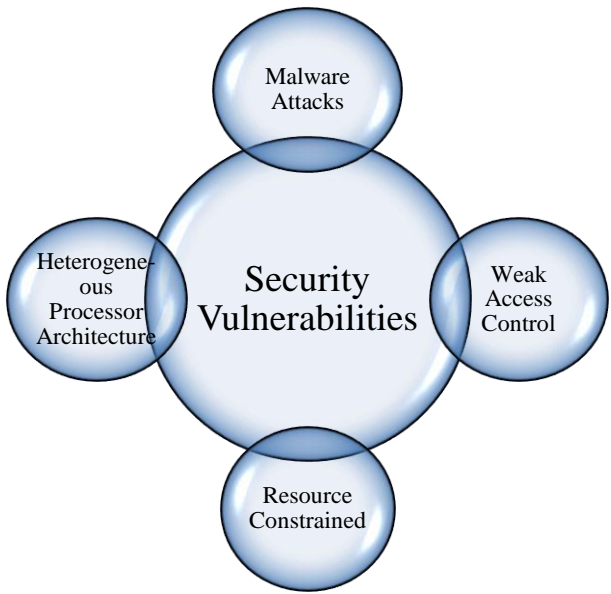

Figure 1: IoT Security Breaches.

To the best of our knowledge, The IOT_Malware detection using deep hybrid and ensemble learning incorporated in this study has not been used in any previous studies. In this study, IoT malware is utilized by its visual image representation and benign files as by the observations, deep CNNs have shown extraordinary performance for the visual representation of challenges [18]. A novel split, transform and merge (S.T.M.) block and squeezed-boosted channel (S.B.) is introduced in the novel model SB-BR-STM for analyzing the feature space and further for malware detection accurately and efficiently in the field of IoT ensemble learning classifiers are used. Additionally, the proposed classification architecture (S.T.M. block of deep CNN) exploits the idea of region-heterogeneity and homogeneity. The main contributions from our side in the current studies are depicted below:

1. A new DSBEL framework is proposed for detecting malware-infected packets in an IoT environment. The framework comprises the stacking of new SB-BR-STM CNN and ensemble classifiers.
2. The novel S.T.M. block and channel-SB ideas are incorporated in the new SB-BR-STM dilated-CNN. Moreover, Max-Pooling, average-pooling, and dilated-convolutional operations are

    incorporated at various levels in the new S.T.M. block for extracting the diverse feature-set, especially consisting of the intensity-homogeneity and heterogeneity, global malicious patterns.

3. TL-based auxiliary-channel extraction and multi-path-based new S.B. idea for achieving various feature map to help improve the proposed SB-BR-STM CNN performance. These operations are employed at initial, mid, and conclusion levels for capturing minute texture variation.

4. The proposed DSBEL framework grants the boosted discriminative features from SB-BR-STM CNN and provides the ensemble classifiers for improving the hybrid learning discrimination.

5. The proposed DSBEL framework's and SB-BR-STM CNN's performance is compared with the existing techniques and evaluated on the IOT_Malware dataset using standard performance measures.

Onwards the paper is presented in the subsections as section 2 represents the related work, our proposed framework is explained in section 3, section 4 discusses the experimental setup, result discussions are given in section 5, and section 6 will conclude the paper.

## 2 Background and Related Work

Malware attacks on IoT devices are growing, and detection using traditional methods is difficult, as these techniques adopted the traditional signatory libraries and interactions expertise of malware analysts. On the other hand, ML and DL techniques can apply to detect malware, which is automatic and adaptable in any discipline [19,20]. ML-based malware detection method involves four steps: construction of the dataset, feature engineering, training of the model, and evaluating the model. Feature engineering calculates the model's validity and characterizes the A.P.K.s by extracting robust and informative features. AndroidManifest.xml file and classes.dex file is the main feature used to characterize the A.P.K.s. Basic information about an A.P.K. is recorded in AndroidManifest.xml, such as requested permissions, hardware information, A.P.K. component, and filtered intents.vClasses. dex is transformed into a small format that consists of Dalvik commands (includes operands and opcode). And disassembling classes.dex files for obtaining advanced features, like flow diagram controlling [21] and API dependency graph [22], can also apply to train the malware detection models. Dynamic behavioral features like network operations, service opening, system calls, file operations, phone calls, and encryption operations can obtain by running the applications on isolated platforms, which has been discussed in the reported literature [23]. These dynamic features, used collectively with the static features, will obtain an exact model and achieve higher detection performance.

Traditional ML models (such as Random Forest [24], Support Vector Machines [31]) and DL (such as CNN [25], Long Short-Term Memory (LSTM) [26,27]) have been extensively used for malware detection. Several ML and DL algorithms provided promising and robust performance for IoT malware detection. These tools employ vulnerability mining in the firmware and applications of IoT, which can infect the whole network or the edge devices of the network [14]. During recent research advancements, an inclination toward ML tools and computational power has increased due to their anti-malware applications. Carrilo et al. [28] used malware detection based on ML under the Linux-based platform malware of IoT by using of data set provided. They also used clustering techniques for malware detection. To detect Mirai botnet attacks in IoTs, Ganesh et al. [29] use ML techniques; they applied the approach of A.A.N. by using of N-BaIot dataset.

Bendiab et al. [18] used the pre-trained ResNet50 for malware traffic analysis in IoT using a 1000 network (pcap) file. A lightweight CNN malware detection approach compared with existing VGG-16 for IoT was reported by Kyushu et al. [21]. In their studies, the central work theme like DDoS malware and IoTPOT used the malware images. Considerable better performance of 95% accuracy were achieved for malware of type DDoS and good ware in their experimental setup [30]. Detection mechanism for android IoT devices, another end-to-end malware mechanism, was introduced by Ren et al. [22] by collecting 8000 malicious and 8000 benign A.P.K. files from virus share and Google play store, respectively. They used significant DL approaches on the Mobile dataset to evaluate their experimental views to detect malware using color images. An active DL-based IIoT malware detection technique has been reported using P.S.E., sparse-autoencoder, and LSTM to train active learners [31]. The fusion framework achieved 95.1% and 86.9% accuracy on detection and adversarial malware detection, respectively. Moreover, the DL-based Bidirectional-Gated Recurrent-Unit-CNN technique has been reported to detect IoT malware and achieved 98% accuracy [11]. All the above-reported work is measured and analyzed in terms of Accuracy and Precision, although the datasets selected are imbalanced. This research work is examined under the benchmark IoT dataset publicly available on Kaggle, and performance evaluation metrics are selected as F1-Score, MCC, and Recall, along with Accuracy and Precision.

## 3 Deep Squeezed-Boosted and Ensemble Learning (DSBEL) Framework

The proposed novel approach comprised of developed a new deep CNN named the Squeezed-Boosted Boundary-Region Split-Transform-Merge SB-BR-STM and ensemble classifiers. The proposed IoT malware detection scheme is comprised of three arrangement schemes: (1) the proposed SB-BR-STM CNN and (2) the DSBEL framework, and (3) evaluating the existing CNNs. The existing customized CNN is used as both learned from scratch and as fine-tuned T.L. using IOT_Malware dataset. Moreover, data augmentation has been performed to improve learning and generalization. Figure 2 is the graphical view of the overall framework.

### 3.1 Data Augmentation

The models of CNN perform better for a large number of labeled data and perform better in generalization. Sometimes, the data points are different from the network requirements. Data augmentation is the process through which the data points are arranged according to the network requirement by image transformations [32], which includes image rotation (0-360 degrees), image scaling (0.5 -1), shearing (-0.5, +0.5), image transformation (grayscale to RGB and vice versa), and reflection (in the right and left direction). Making the data set more robust and generalizing it to a network can be done with the help of the augmentation process.

**3.2 Proposed SB-BR-STM CNN**

In this work, a new deep SB-BR-STM CNN is developed to detect IoT Malware images. The channels are initially systematically split and employ Region-Edge and dilated convolutional operations. Consequently, the channels are squeezed and merged and further fed into fully connected layers. The channels are split into four multi-paths and squeezed to preserve the reduced maps, then boosted after merging for getting diverse feature-maps. The novel channel S.B. approach incorporated at S.T.M. block in a newly modified fashion for capturing minor texture and contrast variation of malicious patterns. The idea of S.B. is employed on the channel at the abstract, mid, and final levels.

Three STM-based blocks are implemented systematically and have the same topology in the proposed SB-BR-STM. Four dilated convolutional blocks constitute the architecture of the SB-BR-STM, as presented by equation 1. Each block applies the average-pooling and max-pooling operations methodically to preserve the region and boundary pattern [33], as organized by equations (2-3). These operations help efficiently assess region homogeneity inside the infected region and determine boundaries, edges, and textural variations.

$$W_{s,t} = \sum_{x=1}^{j} \sum_{y=1}^{k} w_{s+x-1,t+y-1} * u_{x,y} \quad (1)$$

$$W^{avg}_{s,t} = \frac{1}{m^2} \sum_{x=1}^{m} \sum_{y=1}^{m} w_{s+x-1,t+y-1} \quad (2)$$

$$W^{max}_{s,t} = \max_{x=1,\ldots,m,\ y=1,\ldots,m} w_{s+x-1,t+y-1} \quad (3)$$

In equation 1, w represents the input feature map having a dimension of s, t, and u represents the filter having size x, y. the acquired feature vector sorts from the lower level (1) to the upper level (s+x-1) and (t+y-1). The average and max pooling having m size window are represented in Equations (2-3). As depicted in equation 4, channel S.B. operation is improved at every convolutional block (B, C, and D, E) for learning diverse infected feature sets. For attaining diverse feature maps, B & C blocks are produced by TL On the other hand, training from scratch Blocks is D & E. Dimensions of each channel of S.B. convolutional S.T.M. block are 32-128, 64-256, and 128-512, respectively [56]. The main application of T.L. is solving the problem of the target domain by learning the network from the source domain to achieve better performance. Moreover, homogenous operations control distorted regions and outliers in the acquisition of input images [34]. And for achieving optimal features and reducing connection intensity, the boosted channel is processed in block F.

Wd and We are the original blocks of D and E channels, respectively, as shown in equation (4). Similarly, Wb and Wc are depicted as auxiliary block A and C channels generated using T.L., respectively. Then, a(.)

operation will concatenate the original and auxiliary channels. Additionally, dropout layer is used to reduce overfitting and achieve target-specific features. $Z_p$ represents neuron in equation 5. The proposed CNN is represented diagrammatically in Figure 3.

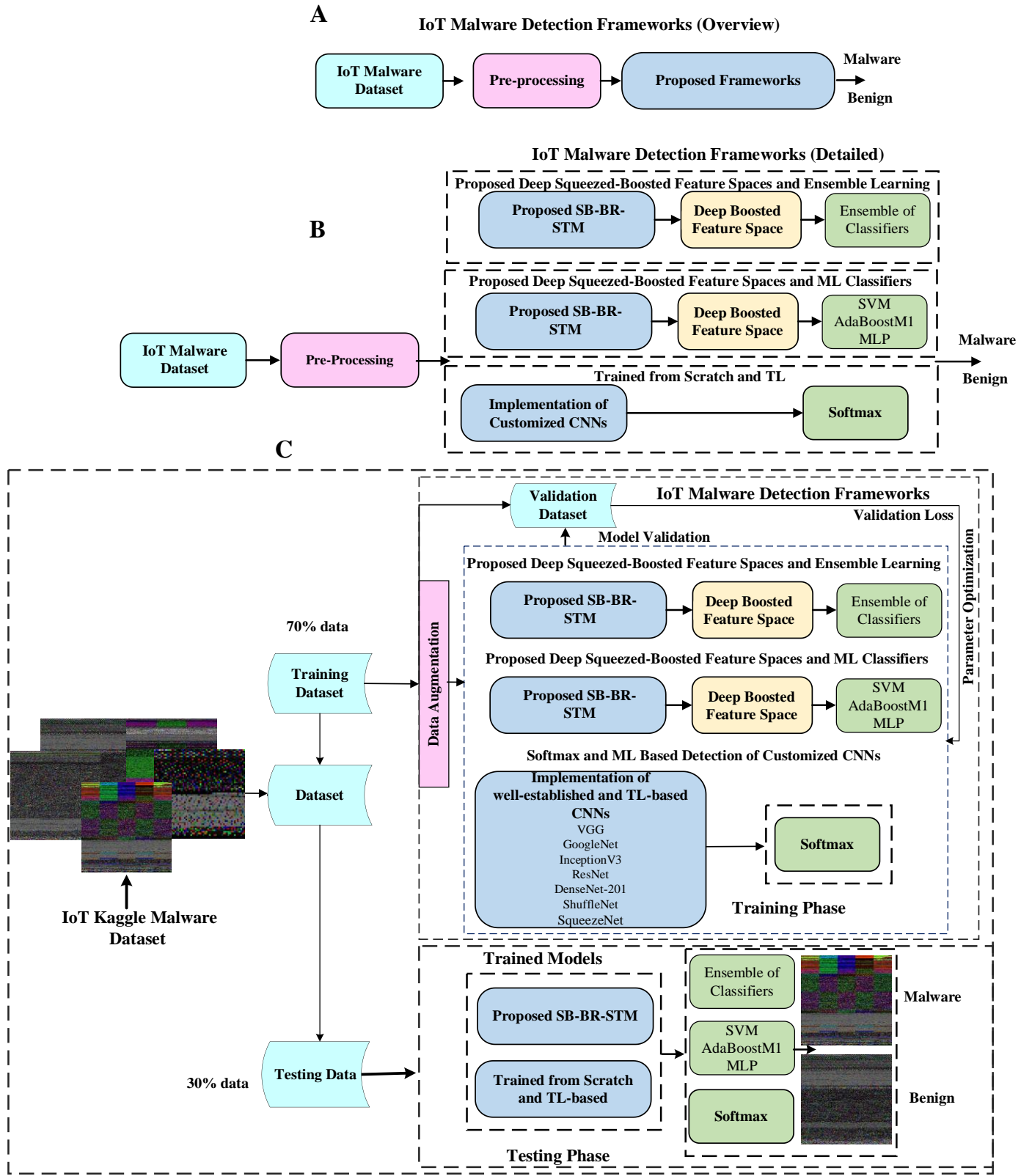

Figure 2: The Proposed Malware Detection Framework.

$$W_{Boosted} = a(w_b || w_c || w_d || w_e) \qquad (4)$$

$$W = \sum_P^p \sum_Q^q z_p w_{Boosted} \qquad (5)$$

$$\sigma(w) = \frac{e^{x_i}}{\sum_{i=1}^c e^{x_c}} \qquad (6)$$

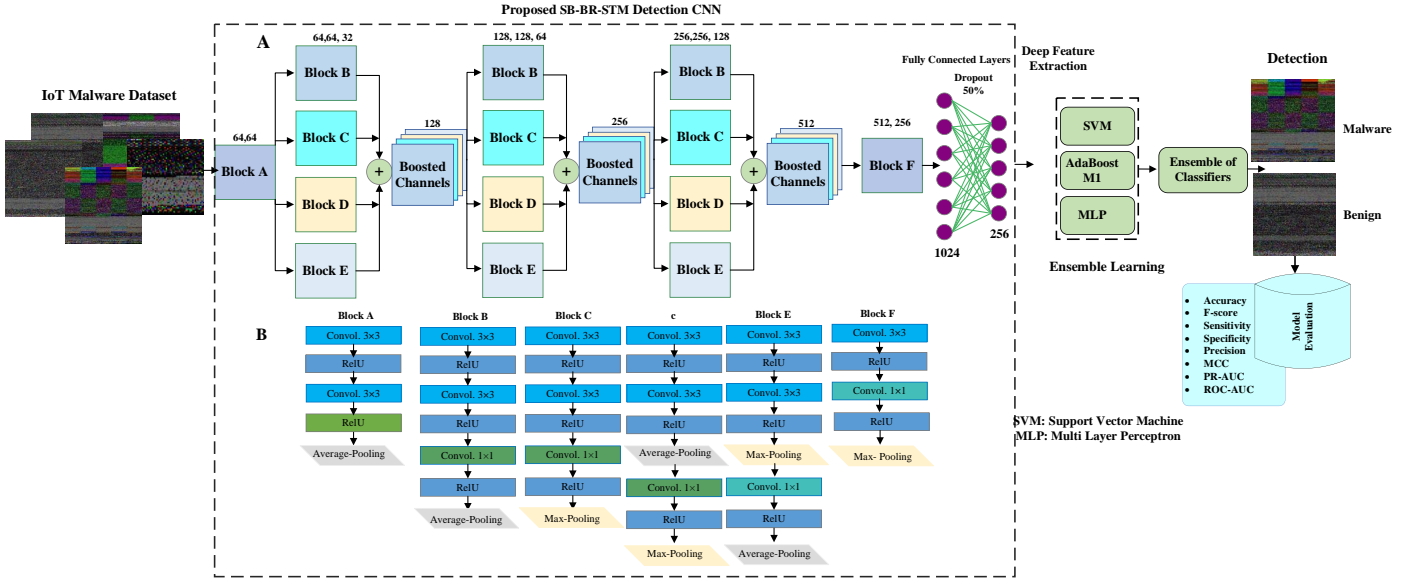

Figure 3: The Proposed DSBEL Framework Comprised of Deep SB-BR-STM CNN.

## 3.3 Significance of Using an Auxiliary Channel

Improvements in the representative capacity of the developed CNN are possible by adding auxiliary channels. These channels are generated using TL-based deep CNN, and for getting prominent feature maps, squeezed at each S.T.M. block and merged at different initial, mid, and high levels. The prominent and diverse channels are learned from different CNN, as shown in the blocks, and concatenated at the next level, which helps in improving image information locally and globally. S.B.-based deep CNN learns complex malicious patterns.

## 3.4 Ensemble ML Classifiers

In the proposed DSBEL framework, Deep-boosted feature space based on the developed SB-BR-STM is fed to ensemble classifiers to detect infected images. The DSBEL framework is applied to acquire a Deep-boosted feature vector and fed into a voting-based ensemble ML classifier. The ensemble ML classifier will take the feature space and detect the infected images by the majority-voting-based method.

The boosted feature vectors are extracted from the proposed SB-BR-STM of diverse channels and fed into the ensemble classifier, as defined in equation (6). Mainly three ML classifiers are used, SVM, M.L.P. [35], and AdaBooSTM1 [36], having the activation functions $f_{SVM}(.)$, $f_{MLP}(.)$, and $f_{AdaBoost}(.)$ as shown in equation (7-10). Optimal hyper-parameters are selected during the training of the proposed CNN, which helps reduce training error and minimizes empirical risk [61]. Moreover, ML classifiers aim to reduce test errors of training set by fixed distribution, exploiting the minimal operational principle and providing generalization.

$$h_{MLP} = f_{MLP}(w) \qquad (7)$$

$$h_{SVM} = f_{SVM}(w) \qquad (8)$$

$$h_{AdaBoost} = f_{AdaBoost}(w) \tag{9}$$

$$h_{final} = f_{Ensemble}(f_{MLP}(w), f_{SVM}(w), f_{AdaBoost}(w)) \tag{10}$$

Encouragement towards combining multiple feature vectors into a single information feature vector and performance improvement aspires from ensemble learning. Also, the unsatisfactory performance risk is avoided by using the extracted feature vectors from a single model. Depending upon the fusion level, applicability can be on classifier ensemble and feature boosting. Boosted feature sets are implicated by the featured ensemble and will feed to the ML classifier to acquire final vectors. On the other hand, the integration of the decision from multiple ML classifiers by the ensemble classifier voting strategy is shown in equation (10). Both features boosting and ensemble classifier techniques have been used in the proposed DSBEL framework.

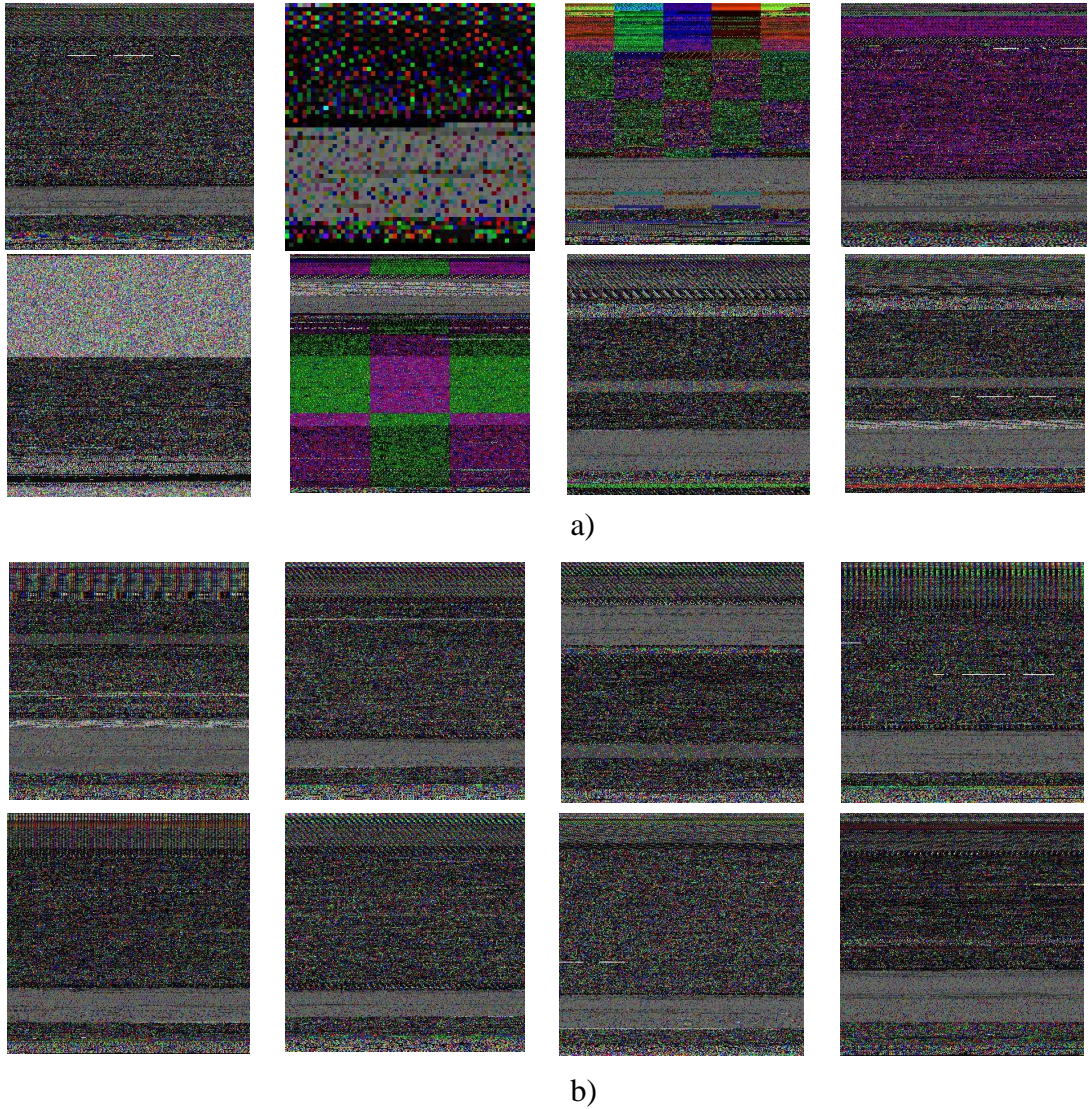

Figure 4: IOT_Malware Visual Representation (a) Malware (b) Benign Ware Images.

### 3.5 Customized CNN Utilization

Many of the current CNN are adopted for detecting malware in IoT and android-based systems [37–39]. The additional layers modify the abstract and final layers of the existing CNN for the requirements of the input and target-class dimension in the dataset. In the train-from-scratch phenomenon, the model learns by back propagating the weights and updating it concerning errors as initial weights are assigned randomly. And on the other hand, in the T.L. techniques, models use pre-trained weights for the convolution layers for convergence improvement. As mentioned, we have also adopted the effectiveness of T.L. for deriving the parameters of the modified prior CNN designed to capture the target-domain characteristics mostly for the malware dataset from the acquired ImageNet trained weights filters.

## 4 Experimental Setup

### 4.1 Dataset

For detecting malware in IoT using deep CNN, all the packets entering the network are represented in the image form because CNN networks are processing the images effectively. The main advantage of image representation is efficiently processing the image and applying it to the trained CNN model to detect malware intrusion. The IOT_Malware dataset is used in this research for training the model [14] IoT-Malware utilizes Byte Sequences of Executable Files [40]. The IOT_Malware dataset consists of two directories: a benign directory containing 2486 images and a malware directory comprising 1473 images. The directory distribution is mentioned in Table 1, and the imagery is represented in Figure 4.

### 4.2 Implementation Detail

The ratio was 70:30% of the data set for training and testing the proposed architecture. Moreover, on the training set, hold-out-validating is performed during the model training at 80:20%. To examine the robustness of the model, this hold-out cross-validation was conducted. All the optimized parameters are selected for validating the model. Table 2 shows the details selected for the model training. MATLAB 2022b using as a tool for developing customized CNNs. NVIDIA GPU GeForce GTX-T dell computer has 32 G.B. of RAM and was used for experimentations. Roundabout 2-4 hours ~ 1-each CNN took 10 minutes/epoch during the training.

Table 1: Benchmark IOT_Malware Dataset Details

| Properties | Description |
|---|---|
| Total | 3,959 images |
| Benign ware | 2,486 images |
| Malware | 1,473 images |
| Train and validation (70%) | (1741, 1032) |
| Test (30 %) | (745, 441) |

### 4.3 Performance Evaluation Metrics

Standard performance metrics were used to evaluate the customized CNNs and proposed SB-BR-STM. These metrics are depicted in Table 3, along with the mathematical explanation. In the classification metrics, Accuracy, F1-Score, and Precision or Sensitivity are included along with True Positive (T.P.), True Negative (T.N.), False Positive (F.P.), and False Negative (F.N.). Equations (7-11) represent Accuracy,

Presicion, Sensitivity, MCC, and F1-Score, respectively. We used Accuracy, Precision, and F1-Score as optimizing metrics for classification in the experimental setup.

Table 2: Hyperparameters for Training CNNs

| Hyper Parameters | Values |
|---|---|
| Learning | $10^{-3}$ |
| Epochs | 20 |
| Optimizer | S.G.D. |
| Batch Dim. | 16 |
| Momentum | 0.950 |

## 5 Results and Discussion

This research presents a novel DSBEL framework and SB-BR-STM CNN for discriminating malware-infected images and benign images in IoT networks. IOT_Malware data set is used to train and validate the proposed network model. This research also evaluates the existing customized state-of-the-art networks and compares the performance with the novel DSBEL framework. Customizing of the existing CNN is incorporated in a modified fashion and implemented using both trains from scratch and Transfer Learning basis. Standard performance measures are used to evaluate the malware detection framework and are shown in Table 4 and Table 5 for both trained from scratch and using T.L., respectively. Table 6 shows the results of the machine learning classifiers and ensemble classifiers.

Table 3: Details of the Assessment Metrics

| Metric Symbol | Description |
|---|---|
| **Accuracy** | Count accuracy in the percentage of the infected points |
| **Precision** | Count how precise the model is, which is the ratio of predicted infected points to the total infected points |
| **Recall / Sensitivity** | Count recall, the proportion of the correctly identified infected points and benign points |
| **MCC** | Mathews Correlation Coefficient |
| **T.P.** | Predicted Correctly Infected Points |
| **TN** | Predicted Correctly Benign Points |
| **F.P.** | Predicted Incorrectly Infected Points |
| **F.N.** | Mispredicted Benign Points |

$$Accuracy = \frac{Predicted\ Infected\ points + Predicted\ Benign\ Points}{Total\ Points} \ X\ 100 \qquad (7)$$

$$Precision = \frac{Predicted\ Infected\ Points}{Predicted\ Infected + Incorrectly\ Predicted\ Infected} \ X\ 100 \qquad (8)$$

$$Sensitivity = \frac{Predicted\ Infected\ Points}{Total\ Infected\ Points} \ X\ 100 \qquad (9)$$

$$MCC = \frac{(TP\ X\ TN) - (FP\ X\ FN)}{\sqrt{(TP+FP)\ X\ (FP+FN)\ X\ (TN+FP)\ X\ (TN+FN)}} \qquad (10)$$

$$F-Score = 2\ X\ \frac{Pre\ X\ Rec}{Pre+Rec} \qquad (11)$$

## 5.1 Performance Analysis of the Proposed SB-BR-STM

Standard imbalance IOT_Malware dataset is used to evaluate the proposed DSBEL performance using standard metrics, Accuracy, F1-score, and MCC. A data augmentation technique was applied to images to increase the learning of the models, and ensemble ML classifiers (SVM, MLP., AdaBooSTM1) are used to detect malicious files, which helped improve the trained model's robustness and generalization. Prediction of the malware in the infected network using the proposed SB-BR-STM showing significant improvements over the traditional networks. In the malware images, textural variation is better explored using the SB-BR-STM by systematically using information related to edge and boundary. Extracting features with different granularity is done using the splitting channel S.B. and merge technique. The TL concepts and S.T.M. incorporated improved the performance of the developed SB-BR-STM compared to the existing CNNs. Significance performance is quantified using the MCC, F1-score, AUC-ROC, Accuracy, Precision, and Recall reported in the current study.

## 5.2 Customized CNN

The existing CNN is customized and compared to its performance with the proposed SB-BR-STM, as shown in Table 4. Training the customized existing models using both training from scratch and Transfer Learning is shown in Table 4 and Table 5. From the tables, it can be better analyzed that the models trained using T.L. perform better than training from scratch. The performance gain of our proposed model and the existing networks using T.L. are (0.52-1.04) % Accuracy, (1.98-7.28) % F1-Score, (0.86-3.69) % MCC, (2.89-10.89) % Sensitivity and (0.36-2.61) % of Precision.

## 5.3 Proposed Boosted and Ensemble Learning Framework

The employed framework is a hybrid learning-based strategy in which the proposed CNN extracts features with the addition of strong ML classifiers. We extracted a deep feature vector from the proposed boosted deep CNN at the end layer and fed it into the ensemble ML competitive classifiers: SVM, M.L.P., and AdaBoostM1. A diverse decision feature space is obtained using the three classifiers and boosted deep feature spaces. Consequently, the boosted features are generated by integrating all these deep features, which maximizes the diversity of feature space, and the discrimination ability of the ML classifier enhances by an ensemble of ML classifiers.

Feature maps are effectively obtained from our proposed model and further fed into the ensemble classifiers (SVM, MLP., and AdabooSTM1) to detect malware in the network packets. Table 6 and Figure 7 shows that selected ML classifiers apply one by one and observe the model performance. After that, a majority of voting-based Ensemble ML classifiers are applied, which shows better performance due to the ensemble technique. The architecture's performance assessment measures Accuracy, Recall, Precision, F-score, and MCC. SB-STM effectiveness is evaluated in the last of the experiment for the proposed SB-BR-STM. Performance gain as (1.01-3.71) % Accuracy, (2.58-7.72) % F1-Score, (2.01-5.43) % MCC, (3.22-12.89) %

Sensitivity, and (0.73-2.22) % of precision is showing the significant improvements of the proposed SB-BR-STM in the malware detection systems, as depicts in Table 4 and Figure 7.

## 5.4 Detection Analysis

Detection and precision rate are the main assessment metrics of the effectiveness of a malware detection architecture. Figure 5 shows the detection performance of existing models and compares them with the model proposed in this research using Accuracy, F1-score, and MCC. The comparison illustrated good Precision by some existing customized models with low Recall. Figure 6 shows the performance gain ranging from minimum to maximum and a comparison with the existing standard CNN architectures. Table 4 and Table 6 depict the summarized results of the proposed SB-BR-STM and DSBEL.

Table 4: Performance Measurements of the Proposed SB-BR-STM. With the Existing Models Using Trained From Scratch

| Models | Accuracy % | F1-Score | MCC | Recall | Precision |
| --- | --- | --- | --- | --- | --- |
| SqueezeNet | 93.47 | 86.76 | 79.14 | 78.26 | 97.34 |
| ShuffleNet | 94.72 | 89.30 | 80.90 | 82.68 | 97.08 |
| Inceptionv3 | 94.89 | 89.11 | 80.91 | 82.28 | 97.17 |
| VGG-16 | 95.38 | 90.14 | 81.29 | 84.29 | 96.86 |
| ResNet-50 | 95.62 | 90.84 | 81.79 | 85.70 | 96.64 |
| GoogleNet | 95.93 | 91.72 | 82.21 | 87.93 | 95.85 |
| DenseNet201 | 96.17 | 91.90 | 82.56 | 87.93 | 96.25 |
| **Proposed SB-BR-STM** | **97.18** | **94.48** | **84.57** | **91.15** | **98.07** |

Table 5: Performance Measurements of SB-BR-STM with the Existing Models Training Using TL

| Models | Accuracy % | F1-Score | MCC | Recall | Precision |
| --- | --- | --- | --- | --- | --- |
| SqueezeNet | 96.14 | 87.20 | 80.88 | 80.26 | 95.46 |
| ShuffleNet | 96.33 | 91.45 | 81.35 | 87.72 | 95.52 |
| Inceptionv3 | 96.47 | 91.71 | 81.95 | 86.80 | 97.20 |
| VGG-16 | 96.58 | 91.31 | 82.51 | 86.23 | 97.02 |
| ResNet-50 | 96.58 | 91.41 | 83.11 | 86.04 | 97.49 |
| GoogleNet | 96.60 | 91.94 | 83.44 | 86.82 | 97.71 |
| DenseNet201 | 96.66 | 92.50 | 83.71 | 88.26 | 97.17 |

Table 6: Performance Analysis of the Proposed Hybrid Frameworks

| Classifier | Accuracy % | F1-Score | Precision | MCC | Recall | AUC |
| --- | --- | --- | --- | --- | --- | --- |
| SVM | 97.71 | 91.87 | 99.80 | 85.88 | 85.11 | 98.83 |
| MLP | 97.79 | 92.72 | 99.61 | 87.11 | 86.72 | 99.18 |
| AdaboostM1 | 97.91 | 99.46 | 99.14 | 89.73 | 90.14 | 99.46 |
| Ensemble (SVM-MLP) | 98.13 | 95.71 | 99.09 | 93.89 | 92.56 | 99.48 |
| **DSBEL (SVM-MLP-AdaboostM1)** | **98.50** | **97.12** | **98.42** | **91.91** | **95.97** | **99.51** |

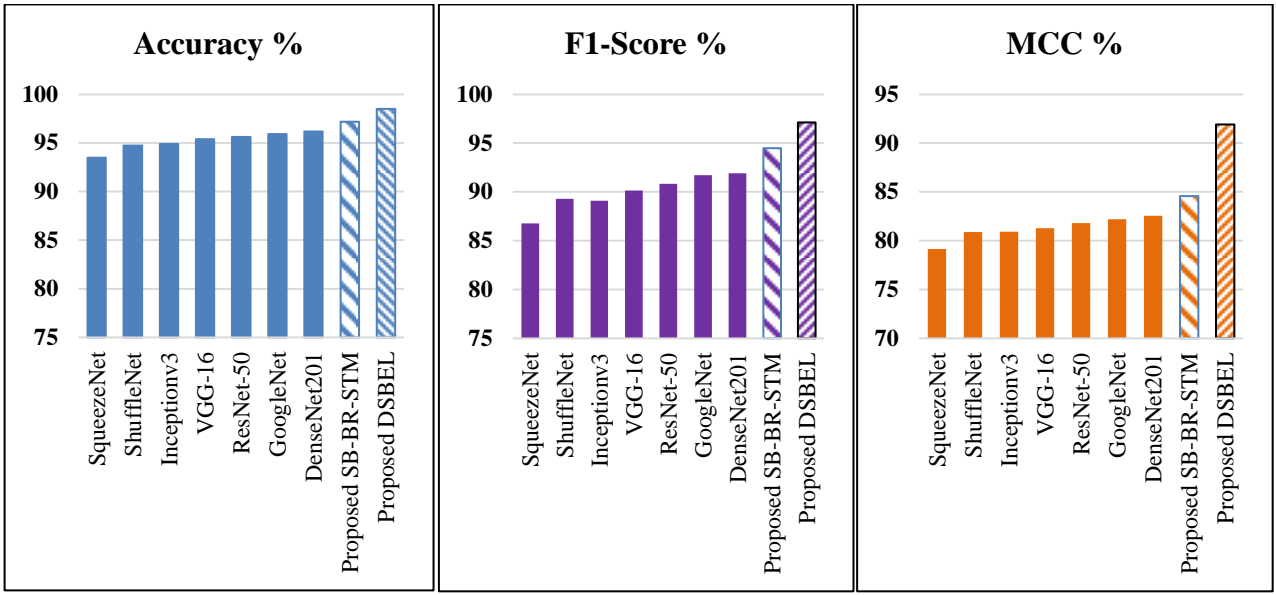

Figure 5: Comparative analysis of existing CNNs VS Proposed SB-BR-STM and DSBEL

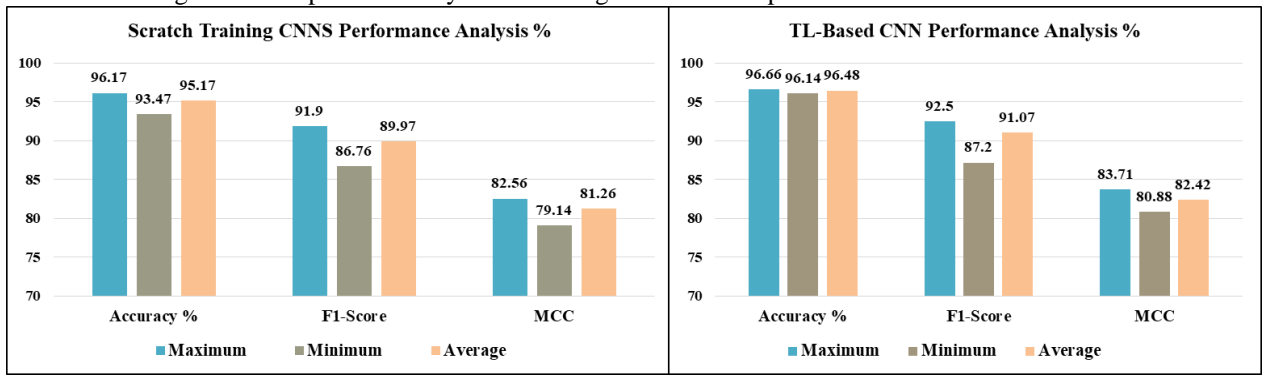

Figure 6: Performance analysis of the existing CNNs.

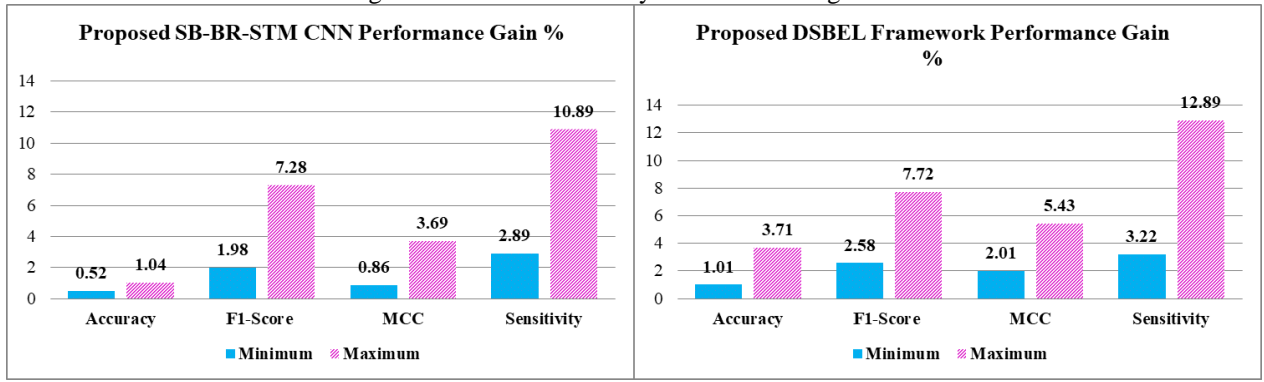

Figure 7: Performance Improvements of the Proposed SB-BR-STM CNN and DSBEL Framework.

## 5.6 Feature Space-Based Analysis

The decision-making of the image is benign or malicious in the proposed model and can be better analyzed through feature space visualization. Prominent visual features and better discrimination factors of the models are associated, which helps to lower the variance and improves the learning rate of the model. Figure 8 and Figure 9 show the proposed SB-BR-STM and DSBEL principal components of feature space visualization. IoT malware features are extracted at different levels by channel squeezed and boosted techniques in S.T.M. blocks. Moreover, by incorporating channel concatenation, S.T.M. boosted the reduced prominent features. An upgrade in identifying distinct and diverse features is shown by visualization for the developed DSBEL and improved detection rate of the IoT malware files.

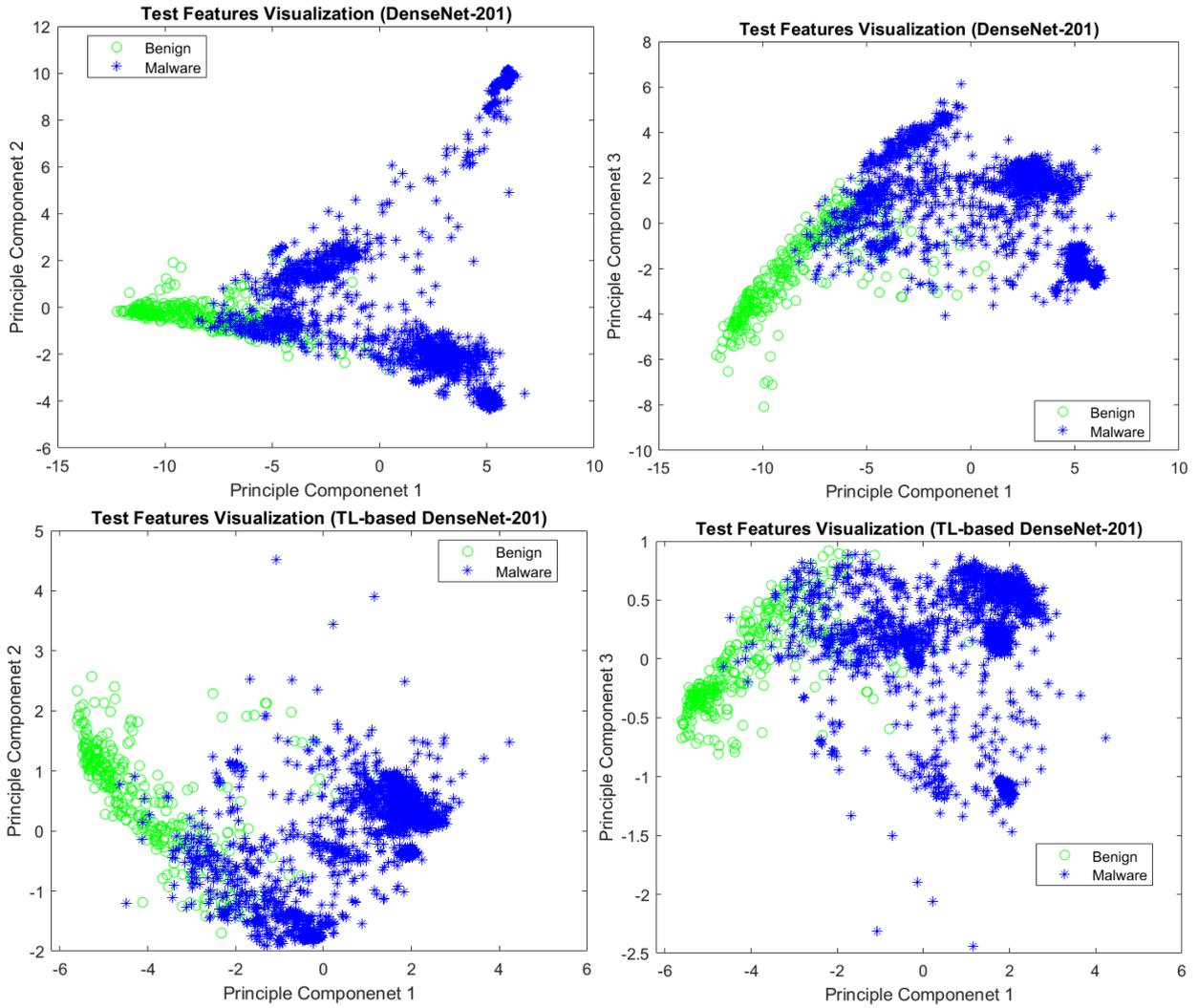
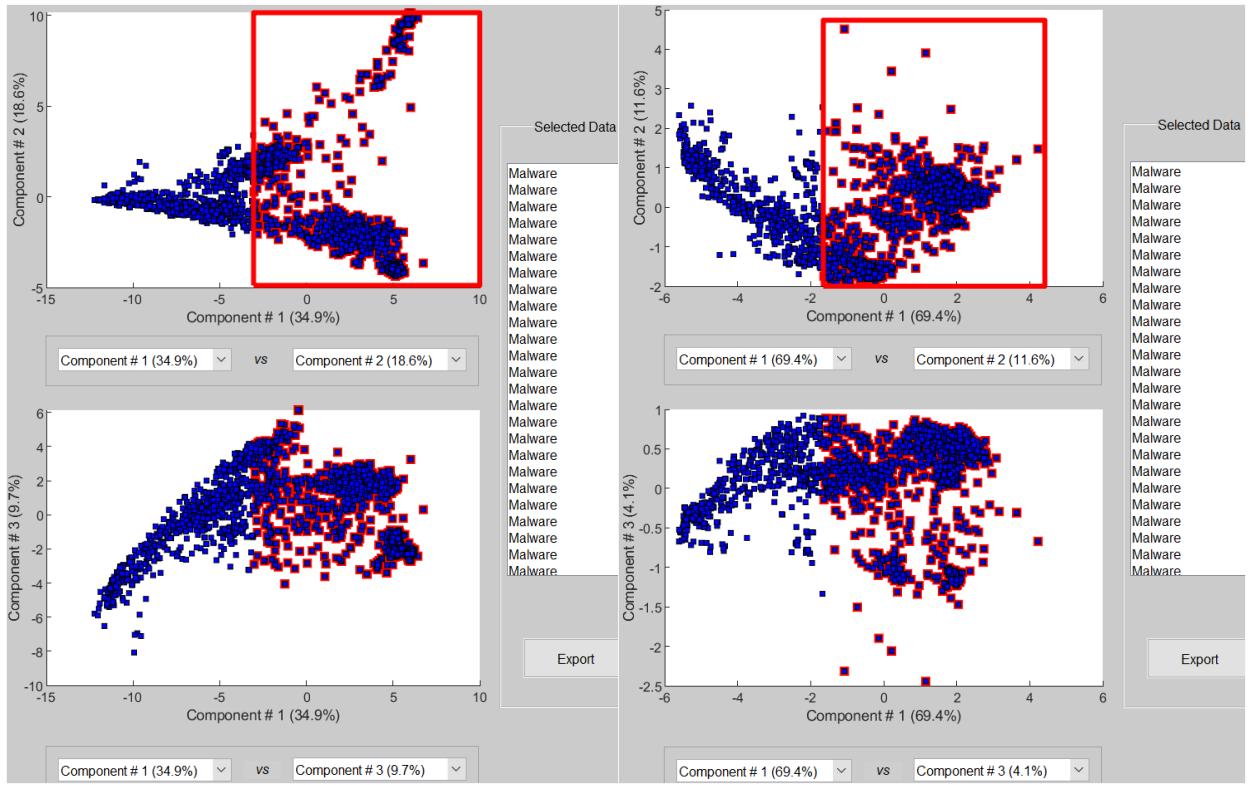

Figure 8: Feature visualization PC1, PC2, and PC3 on the best Performing in Existing CNN (DenseNet-201).

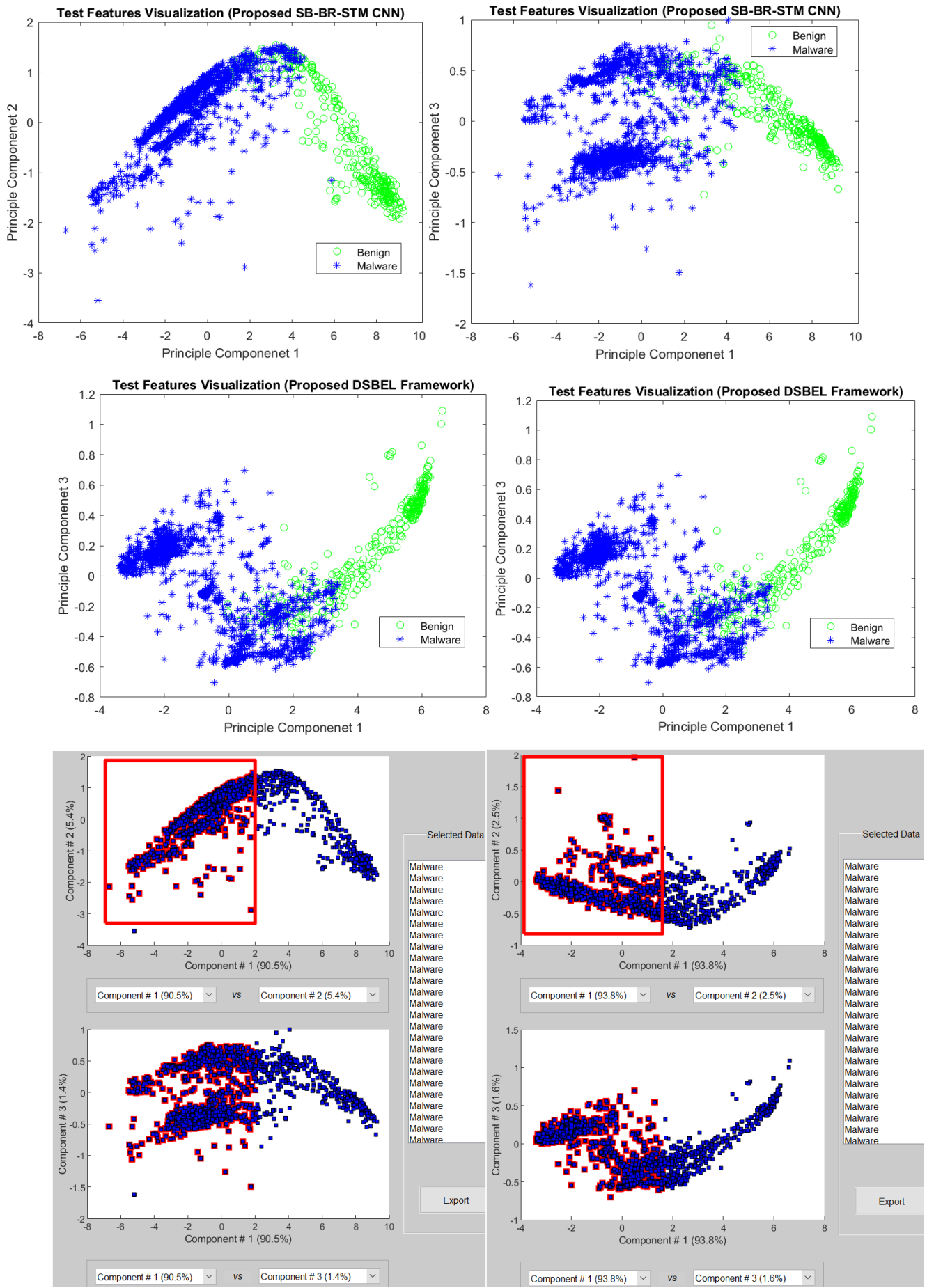

Figure 9: Feature visualization PC1, PC2, and PC3 for the Proposed DSBEL and SB-BR-STM CNN.

## 5.7 ROC and PR-curve Analysis

These are the graphical representation of the classifier's capability of discriminating at all the possible values dimension. The optimum detection cut-off of a classifier can significantly access by the ROC curve [41]. Figure 10 shows the visualized results using the ROC curve with malware images as positive class detection organized by AUC. High values of AUC show the low false positive of the proposed framework with greater sensitivity and considerable performance in the filtration of the malware-infected network.

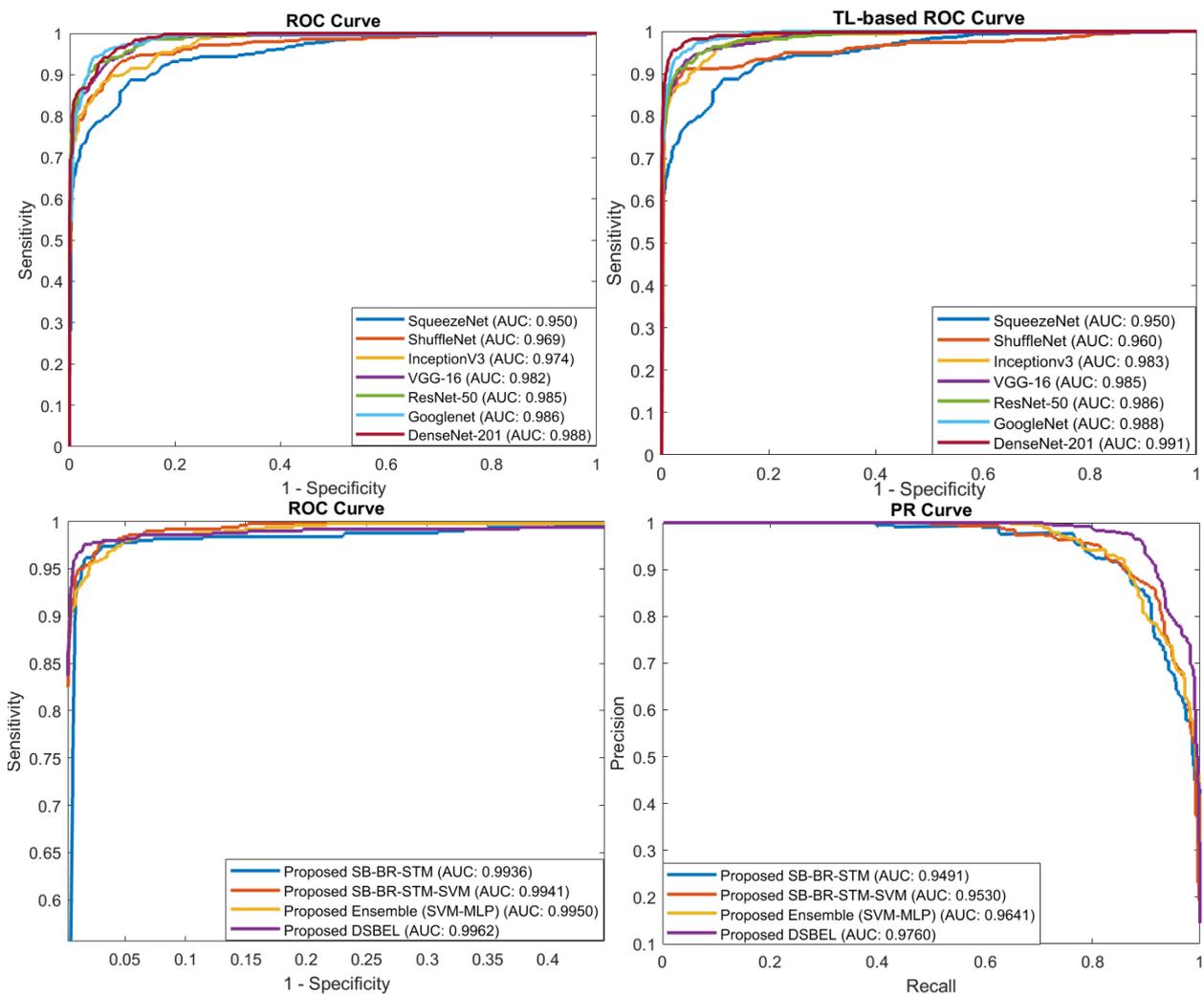

Figure 10: ROC and PR curves for the Developed SB-BR-STM CNN and DSBEL Framework contrasts Existing CNNs.

## 6 Conclusion

IoT Malware detection is useful for the timely detection of malicious activity and suggests methods for future early detection and mitigation. In this regard, we have introduced a deep DSBEL framework that assembles the developed SB-BR-STM's boosted features and ensemble classifiers to detect malware-attacked packets in the IoT network. The SB-BR-STM comprises the S.T.M. blocks that employ TL-based

SB ideas and global-boundary and local-regional operations to preserve diverse and boosted features. Moreover, ensemble learning is used to detect malware patterns based on the obtained features from SB-BR-STM for better discrimination and generalization of the DSBEL framework. The proposed novel hybrid framework is empirically assessed and shows prominent performance with an Accuracy of 98.50%, F1-Score of 97.12%, Recall of 95.97 %, and Precision of 98.42 %. The proposed DSBEL framework may be proficient enough to find attacks of cross platforms malware and stringent environments. The malware includes certain similarities in either forum, and these similar features can help in their identification and detection. In the future, online and android malware detection can be performed with the help of a robust DSBEL framework for real-time realization.

**Acknowledgments**

We thank the Department of Computer Systems Engineering, the University of Engineering and Applied Sciences (UEAS), Swat, Pakistan, for providing the necessary computational resources and a healthy research environment.